\documentclass{article}
\usepackage{ijcai20}

\usepackage{times}

\usepackage{soul}
\usepackage{url}
\usepackage[hidelinks]{hyperref}
\usepackage[utf8]{inputenc}
\usepackage[small]{caption}
\usepackage{graphicx}
\usepackage{amsmath}
\usepackage{booktabs}
\usepackage{multirow}
\usepackage{makecell}
\usepackage{amssymb}
\usepackage{xspace}
\makeatletter
\DeclareRobustCommand\onedot{\futurelet\@let@token\@onedot}
\def\@onedot{\ifx\@let@token.\else.\null\fi\xspace}
\def\eg{\emph{e.g}\onedot} 
\def\ie{\emph{i.e}\onedot} 
 
\def\etc{\emph{etc}\onedot}

\makeatother

\title{Non-Autoregressive Image Captioning \\ with 
	Counterfactuals-Critical Multi-Agent Learning}

\author{
Longteng Guo$^{1,2}$\and
Jing Liu$^1$\footnote{Corresponding Author}\and
Xinxin Zhu$^1$\and
Xingjian He$^{1,2}$\and
Jie Jiang$^{1,2}$\And
Hanqing Lu$^1$\\
\affiliations
$^1$National Laboratory of Pattern Recognition, Institute of Automation, Chinese Academy of Sciences\\
$^2$School of Artificial Intelligence, University of Chinese Academy of Sciences\\
\emails
\{longteng.guo, jliu, xinxin.zhu, xingjian.he, jie.jiang, luhq\}@nlpr.ia.ac.cn }

\begin{document}

\maketitle

\begin{abstract}
	Most image captioning models are autoregressive, \ie they generate each word by conditioning on previously generated words, which leads to heavy latency during inference. Recently, non-autoregressive decoding has been proposed in machine translation to speed up the inference time by generating all words in parallel. Typically, these models use the word-level cross-entropy loss to optimize each word independently. However, such a learning process fails to consider the sentence-level consistency, thus resulting in inferior generation quality of these non-autoregressive models. In this paper, we propose a Non-Autoregressive Image Captioning (NAIC) model with a novel training paradigm: Counterfactuals-critical Multi-Agent Learning (CMAL). CMAL formulates NAIC as a multi-agent reinforcement learning system where positions in the target sequence are viewed as agents that learn to cooperatively maximize a sentence-level reward. Besides, we propose to utilize massive unlabeled images to boost captioning performance. Extensive experiments on MSCOCO image captioning benchmark show that our NAIC model achieves a performance comparable to state-of-the-art autoregressive models, while brings $13.9\times$ decoding speedup. 
\end{abstract}

\section{Introduction}
Image captioning \cite{vinyals2017show,guo2019mscap} aims at generating a natural language description of an image. 
Recent methods typically follow the 
encoder/decoder paradigm where a convolutional neural network (CNN) encodes the input image, 
and a sequence decoder, \eg recurrent neural networks (RNNs) or Transformer \cite{vaswani2017attention}, generates a caption. 
Most of these models use \textit{autoregressive} decoders that require sequential execution: %
they generate each word conditioned on the sequence of previously generated words. 
However, this process is not parallelizable and thus results in high inference latency, 
which is sometimes unaffordable for real-time industrial applications. 

\begin{figure}[!t] 
	\centering
	\includegraphics[width=3.3in]{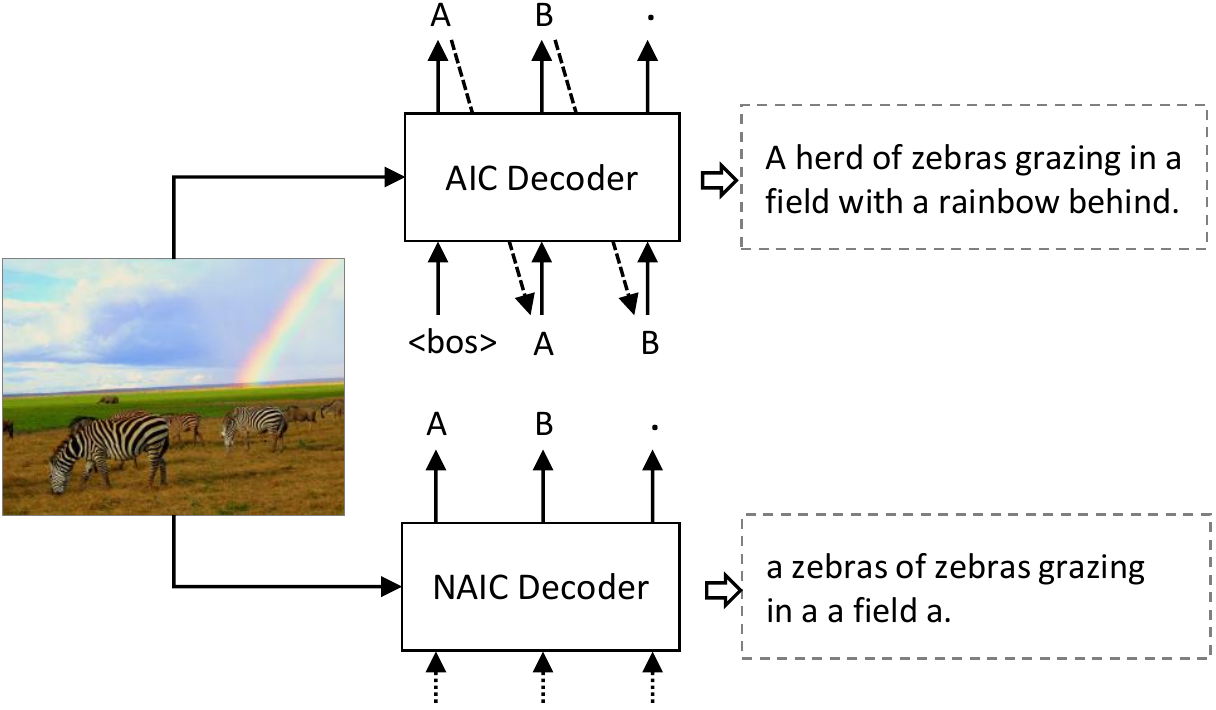}
	\caption{
		Given an image, autoregressive image captioning (AIC) model generates a caption word by word, while 
		Non-Autoregressive Image Captioning (NAIC) model outputs all words in parallel. 
	}
	\label{fig:first}
\end{figure}

Recently, {\em non-autoregressive} decoding was proposed in neural machine translation (NMT) \cite{gu2017non} to significantly improve
the inference speed by predicting all target words in parallel.  
A non-autoregressive model takes basically the same structure as the autoregressive Transformer network \cite{vaswani2017attention}.  
But instead of conditioning the decoder on the previously generated words as in autoregressive models, 
they generate all words independently, as is illustrated in Figure~\ref{fig:first}. 
Such models are typically optimized by the cross-entropy (XE) losses of individual words. 

However, existing non-autoregressive models still have a large gap in generation quality compared to their autoregressive counterparts, mainly due to their severe decoding inconsistency problem. 
For example, in Figure~\ref{fig:first}, the caption generated by the non-autoregressive model has repeated words and incomplete content. 
A major reason for such performance degradation is 
that the word-level XE based training objective cannot guarantee the sentence-level consistency. 
That is, the XE loss encourages the model to generate the golden word in each position 
without considering the global consistency of the whole sentence. 

To simultaneously reduce the inference time and improve the decoding consistency of image captioning, 
in this paper, we propose a Non-Autoregressive Image Captioning (NAIC) model with a novel training paradigm: 
Counterfactuals-critical Multi-Agent Learning (CMAL). 
Specifically, we consider NAIC as a cooperative multi-agent reinforcement learning (MARL)  \cite{bucsoniu2010multi} system, 
where positions in the target sequence are viewed as ``agents" that act cooperatively to maximize the quality of the whole sentence. 
Each agent observes the ``environment" (encoded visual context), 
and communicates with other agents through the self-attention layers in Transformer. 
After several rounds of environment observation and agent communication, 
the agents reach an agreement about content of the target sentence and separately take ``actions" to predict the words in their corresponding positions. 
The agents then receive a common sentence-level reward 
and use policy gradient to update their parameters. 
A benefit of this training paradigm is that the non-differentiable test metrics of image captioning could be directly optimized. 
Another benefit is that by optimizing the agents towards a common sentence-level objective, the decoding consistency can be substantially improved.

A crucial challenge in the above MARL training paradigm is multi-agent credit assignment \cite{chang2004all}: 
the shared team-reward making it difficult for each agent to deduce its own contribution to the team’s success. 
This could impede multi-agent learning and lead to decoding inconsistency. 
To address this challenge, we compute an agent-specific advantage function that compares the team-reward 
for the joint action against an agent-wise \textit{counterfactual baseline} \cite{foerster2018counterfactual,chen2019counterfactual}. 
The counterfactual baseline of an agent is the expected reward when marginalizing out a single agent’s action, while keeping the other agents’ actions fixed. 
As a result, only actions from an agent that outperform the counterfactual baseline are given positive weight, 
and inferior actions are suppressed. 
CMAL fully exploits the distinctive features of the multi-agent NAIC system: extremely short episode and large action space.

To further boost captioning performance, we propose to utilize massive unlabeled images as additional data for training, 
which could be more easily obtained without costly human annotations. 
We evaluate the proposed method on the challenging MSCOCO \cite{chen2015microsoft} image captioning benchmark. 
Experimental results show that our method brings $13.9\times$ decoding speedup relative to the autoregressive counterpart, 
while achieving comparable performance to state-of-the-art autoregressive models.

To summarize, the main contributions of this paper are three-fold: 

\begin{itemize}
	\item  We propose a Non-Autoregressive Image Captioning (NAIC) model with a novel training paradigm: 
	Counterfactuals-Critical Multi-Agent Learning. 
	To the best of our knowledge, we are the first to formulate non-autoregressive sequence generation 
	as a cooperative multi-agent problem. 
	\item 
	We design a counterfactual baseline to disentangle the individual contribution of each agent from the team-reward.
	\item We propose to utilize massive unlabeled data to boost the performance of non-autoregressive models. 
	\item Our method significantly improves the inference speed of image captioning, while at the same time achieves a  
	performance comparable to state-of-the-art autoregressive image captioning methods. 
\end{itemize}

\section{Related Work}
\paragraph{Non-Autoregressive Sequence Generation.}
Non-Autoregressive neural machine Translation (NAT) \cite{gu2017non} has recently been introduced 
to speed up the inference process for real-time decoding, but often performs worse than the autoregressive counterparts. 
Some methods has been proposed to narrow the performance gap between autoregressive and non-autoregressive models, 
including knowledge distillation \cite{gu2017non}, auxiliary regularization terms \cite{wang2019non}, 
well-designed decoder input \cite{guo2019non}, 
iterative refinement \cite{lee2018deterministic,gao2019masked} \etc. %
Among them, MNIC \cite{gao2019masked} and FNIC \cite{fei2019fast} are published works on non-autoregressive image captioning. 
However, these methods are trained with conventional XE loss, 
which is not sentence-level consistent. 
Unlike these works, we propose using CMAL to optimize a sentence-level objective.

\paragraph{Multi-Agent Reinforcement Learning (MARL).} 
MARL \cite{bucsoniu2010multi} considers a system of agents that interact within a common environment. 
It is often designed to deal with complex reinforcement learning problems that require decentralised policies, where each agent selects its own action. 
Compared to well-studied MARL game tasks, our NAIC model has a much larger action space   
and extremely shorter episode.  
Our counterfactual baseline gets intuition from \cite{foerster2018counterfactual}, which 
requires training an additional critic network to estimate the Q value for each possible action. 
Learning such a critic network increases the model complexity and is not practical due to the high-dimensional action space of NAIC. 
Instead, following \cite{chen2019counterfactual}, we turn to the simple yet powerful REINFORCE \cite{Williams1992Simple} algorithm that directly uses the actual return to replace Q function. 

\section{Background}
\subsection{Autoregressive Decoding}
Given an image $I$ as input and a target sentence $y=(y_1,..., y_T)$, AIC models 
are based on a chain of conditional probabilities 
with a left-to-right causal structure: 
\begin{equation}
p(y | I; \theta)=\prod_{i=1}^{T} p\left(y_{i} | y_{<i}, I; \theta \right),
\end{equation}
where $\theta$ is the model's parameters and $y_{<i}$ represents the words before the $i$-th word of target $y$. 
The inference process is not parallelizable under such autoregressive factorization as 
the sentence is generated word by word sequentially. 

\subsection{Non-Autoregressive Decoding}
Recently, non-autoregressive sequence models were proposed to alleviate the inference 
latency by removing the sequential dependencies within the target sentence. 
A NAIC model generates all words independently:
\begin{equation}
p(y | I; \theta)=\prod_{i=1}^{T} p\left(y_{i} | I; \theta \right). 
\label{eqn:na}
\end{equation}
During inference, all words could be parallelly decoded in one pass, 
thus the inference speed could be significantly improved. 

\paragraph{Maximum Likelihood Training.}
Typically, a non-autoregressive sequence model straightforwardly adopts maximum likelihood training with a cross-entropy (XE) loss applied
at each decoding position $i$ of the sentence: 
\begin{equation}
\mathcal{L}_{XE}(\theta)=-\sum_{i=1}^{T} \log \left(p\left(y_{i} | I; \theta \right)\right)
\label{eqn:xe}
\end{equation}

\begin{figure}[!t] 
	\centering
	\resizebox{0.51\textwidth}{!}{
		\includegraphics[width=6in]{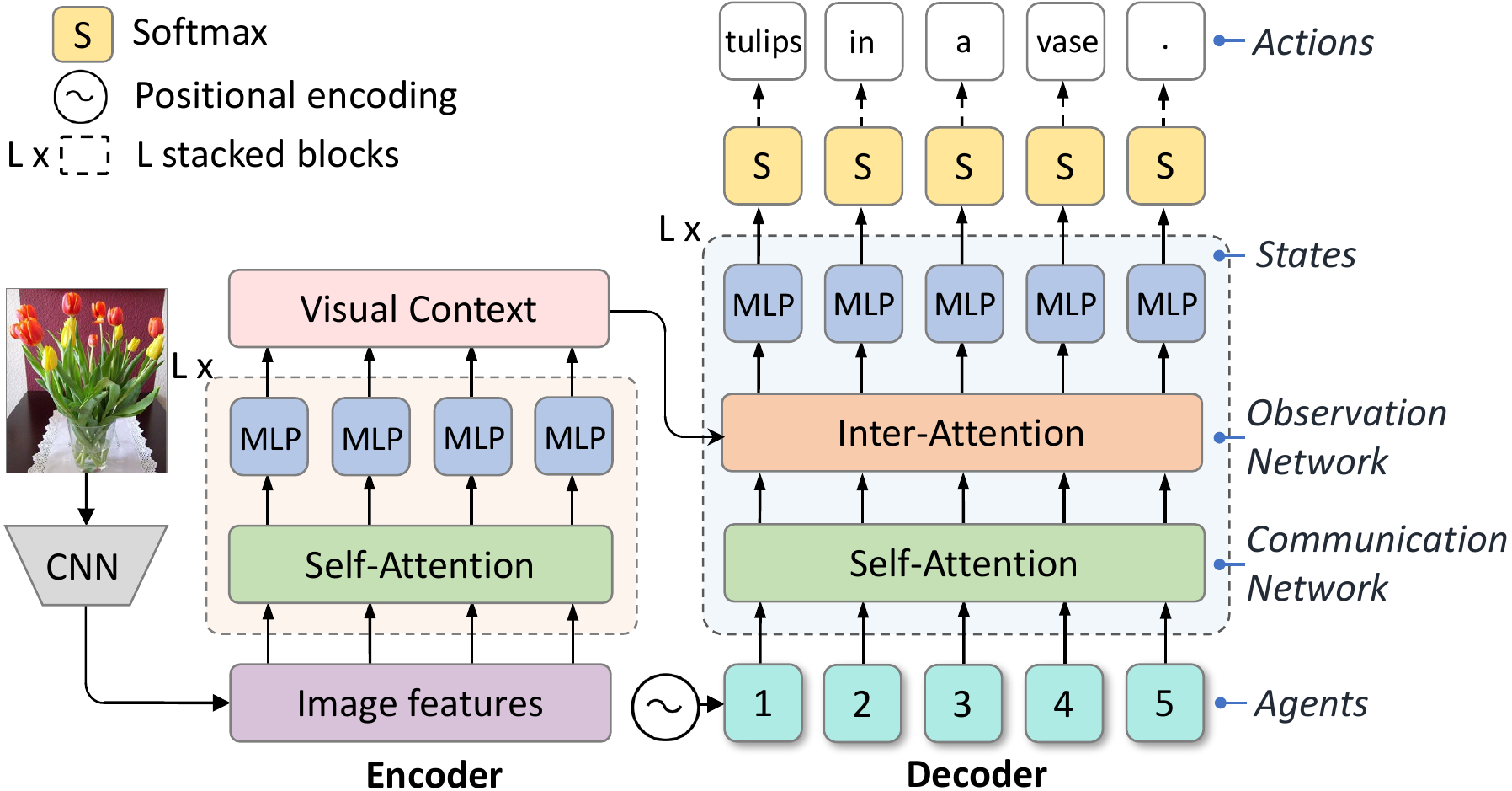}
	}
	\caption{
		Illustration of our Transformer-based non-autoregressive image captioning model, which composes of an encoder and a decoder. 
		On the rightmost, we cast the non-autoregressive decoder in the multi-agent reinforcement learning terminology. 
	}
	\label{fig:framework}
\end{figure}

\section{Approach} 
In this section, we first present the architecture of our NAIC model, 
and then introduce our Counterfactuals-critical Multi-Agent Learning (CMAL) algorithm for model optimization. 
Finally, we describe how we utilize unlabeled data to boost captioning performance. 
\subsection{Transformer-Based NAIC Model} 
Given the image features of an image, 
NAIC generates a caption about that image in a non-autoregressive manner. 
The architecture of our NAIC model is based on the well-known Transformer network,  
which composes of an encoder and decoder, as is shown in Figure~\ref{fig:framework}.

\paragraph{Image features and encoder.} 
Following previous works on image captioning \cite{anderson2017bottom}, given an image, we first extract vectorial image features from a pre-trained CNN network. 
The encoder of NAIC is basically the same as the Transformer encoder, which takes the image features as inputs and generates the visual context representation. 

\paragraph{Decoder.} 
Since the sequential dependency is removed in the non-autoregressive decoder, previous works often introduce additional components \eg 
target length predictor, well-designed decoder architecture and decoder inputs \etc, which adds on extra inference time. 
Different from these works, we choose a design that simplifies the decoder to the most degree but proves to work well in our experiment. 
That is, we keep the decoder architecture almost the same as the Transformer decoder, 
and simply use a sequence of sinusoidal positional encodings \cite{vaswani2017attention} as the decoder input, 
each of which represents a position in the target sequence. 
We remove the autoregressive mask from the self-attention layers of the decoder, 
allowing each position in the decoder to attend over all positions in the decoder.

\subsection{Counterfactuals-Critical Multi-agent Learning} 
\subsubsection{NAIC as a MARL Problem}
To address the decoding inconsistency problem caused by word-level XE loss and directly optimize non-differential test metrics, 
we formulate NAIC model as a fully cooperative Multi-Agent Reinforcement Learning (MARL) system. 
We now formally cast NAIC in the MARL terminology.

\textbf{Agent.} Each word position in the target sequence is viewed as an agent that interacts with
a common ``environment" (visual context from the encoder output) and other agents. There are $N$ agents in total, 
identified by $a \in A \equiv\{1, \ldots, N\} $. 

\textbf{State.} The hidden states in our NAIC decoder layers naturally represent the states of the agents, 
which are updated in each decoder layer. 
The agents observe the ``environment" through the inter-attention layer where they attend to the visual context, 
and communicate with other agents through the self-attention layer where the messages are passed between every two agents. 
After $L$ rounds of observation and communication, the final state of each agent is denoted as $s_a$. 

\textbf{Action.} After obtaining $s_a$, each agent simultaneously chooses an action $u_a \in U$, 
which is a word from the whole vocabulary $U$.
The actions of all agents form a joint action $\mathbf{u} \in \mathbf{U} \equiv U^{N}$. %
To transform the joint action into a sentence, we truncate the word sequence at the first period. 

\textbf{Policy.} The parameters of the network, $\theta$, define a stochastic policy $\pi_a$ for each agent, 
from which the action is sampled, \ie $u_a \sim \pi_a = \textit{softmax}(s_a)$. 
We speed learning and reduce model complexity by sharing parameters among agents. 

\textbf{Reward.} 
After all agents take their actions (words),  
they receive a shared ``team-reward" $R(\mathbf{u})$.  
The reward is computed with an evaluation metric (\eg CIDEr) 
by comparing the generated sentence to corresponding ground-truth sequences.

Compared to typical MARL applications, NAIC has a much larger action space (\ie the whole vocabulary, which is near 10,000 words),  
and extremely shorter episode (\ie the episode length is 1).  
Actually, agents in NAIC perform a one-step Markov Decision Process (MDP) since all words are generated in one-pass. 
We denote joint quantities over all agents in bold, \eg $\mathbf{u}$, $\boldsymbol{\pi}$.

\subsubsection{Multi-Agent Policy Gradient} 
The goal of multi-agent learning is to maximize the expected team-reward. 
With the policy gradient theorem, the expected gradient of the agents can be computed as follows: 
\begin{equation}
\nabla_{\theta} \mathcal{L}(\theta)=-\mathbb{E}_{\boldsymbol{\pi}} \left[ \sum_a R(\mathbf{u})  \nabla_{\theta} \log \pi_a \left(u_a|s_a; \theta \right)\right]. 
\end{equation} 
Particularly, using the REINFORCE \cite{Williams1992Simple} algorithm, 
the above equation can be approximated using a single sampling $\mathbf{u} \sim \boldsymbol{\pi} $ from the agents: 
\begin{equation}
\nabla_{\theta} \mathcal{L}(\theta) \approx  \sum_a R(\mathbf{u})  \nabla_{\theta} \log \pi_a \left(u_a|s_a; \theta \right) . 
\end{equation}

However, such a gradient estimate suffers from high variance, 
which leads to unstable and slow learning of the optimal policy. 
To reduce the variance, a reference reward or \textit{baseline} $b$ can be subtracted from the reward: 
\begin{equation}
\nabla_{\theta} \mathcal{L}(\theta) \approx  \sum_a (R(\mathbf{u}) - b)  \nabla_{\theta} \log \pi_a \left(u_a|s_a; \theta \right) . 
\label{eqn:pg}
\end{equation}
The baseline still leads to an unbiased estimate, and importantly, it results in lower variance of the 
gradient estimate \cite{sutton1998reinforcement}. 
The baseline can be an arbitrary function, as long as it does not depend on the action $u_a$.

\subsubsection{Counterfactual Baseline}

The above approach, however, fails to address a key multi-agent credit assignment problem. 
That is, because each agent receives the same team-reward, 
it is unclear how a specific agent’s action contributes to that team-reward. 
The consequences of this problem are inefficient multi-agent learning and decoding inconsistency.  
For example, suppose there is a generated sentence (joint action), ``a girl \underline{girl} riding a bike", and it gets a relatively high reward, 
then the word ``girl" taken by the 3rd agent is likely to be pushed up because it receives a positive reward, 
which, however, should actually be suppressed and replaced with ``is". 

To address this problem, we decide to compute a separate advantage function $A_a(s_a, \mathbf{u})$ for each agent. 
It is computed by 
subtracting an agent-specific \textit{counterfactual baseline} $B_a( s_a, \mathbf{u}_{-a} )$ from the common team-reward, \ie:
\begin{equation}
A_a(s_a, \mathbf{u}) = R(\mathbf{u}) - B_a( s_a, \mathbf{u}_{-a} ), 
\end{equation}
where $\mathbf{u}_{-a}$ denotes the joint action of all the agents other than agent $a$. 
$A_a(s_a, \mathbf{u})$ measures the increase (or decrease) in expected return of a joint action $\mathbf{u}$ due to agent $a$ having chosen action $u_a$ under state $s_a$. 
The gradient in Equation \ref{eqn:pg} then becomes:
\begin{equation}
\nabla_{\theta} \mathcal{L}(\theta) \approx  \sum_a A_a(s_a, \mathbf{u})  \nabla_{\theta} \log \pi_a \left(u_a|s_a; \theta \right) . 
\label{eqn:pg-cf}
\end{equation}
Since $B_a( s_a, \mathbf{u}_{-a} )$ does not depend on the action of agent $a$,  
as described above, it will not change the expected gradient. 

Formally, the counterfactual baseline $B_a$ is calculated by marginalizing the rewards when agent $a$ traverses all possible actions while keeping the other agents' actions $\mathbf{u}_{-a}$ fixed:

\begin{equation}
B_a( s_a, \mathbf{u}_{-a} ) = \mathbb{E}_{u^\prime_a \sim \pi_a} \left[ R([\mathbf{u}_{-a}, u^\prime_a]) \right].
\label{eqn:expect}
\end{equation}

The key insight of using this counterfactual baseline for NAIC is that: 
given a sampled sequence/joint-action, if we replace the chosen word/action of a target position/agent with all possible words/actions and see how such counterfactual replacements affect the resulting reward, then the expected reward can act as a baseline to tell the actual influence of the chosen word/action. 
As a result, for each agent, only actions that outperform its counterfactual baseline would be pushed up, 
and inferior actions would be suppressed. 

Because the action space of each agent is quite large, 
we approximate the expectation computation in the above equation by only considering $k$ actions with the highest probability: 
\begin{equation}
\begin{aligned}
B_a( s_a, \mathbf{u}_{-a} ) &\approx \sum_{u^\prime_a \in \mathcal{T}_{a}} \pi_a^\prime \left(u^\prime_a|s_a; \theta \right)  R([\mathbf{u}_{-a}, u^\prime_a]), \\
\pi_a^\prime \left(u_a|s_a; \theta \right)& = \frac{ \pi_a \left(u_a|s_a; \theta \right) }{ \sum_{u^\prime_a \in \mathcal{T}_{a}} \pi_a \left(u^\prime_a|s_a; \theta \right) },
\label{eqn:topk}
\end{aligned}
\end{equation}
where $\mathcal{T}_{a}$ is the set of words with top-$k$ probabilities in $\pi_a$, 
and $\pi_a^\prime \left(u_a|s_a; \theta \right)$ is the re-normalized probability for action $u_a$. 
Experimentally, we found this approximation to be quite accurate even with a relatively small $k$ 
because the top-ranking words often have dominating probabilities. 

Thanks to the one-step MDP nature of our NAIC model, 
the counterfactual replacements could be effortlessly made 
by simply choosing new words from $ \pi_a \left(u_a|s_a; \theta \right)$, without the need for time-consuming Monte-Carlo rollouts as in common multi-step MDP problems.

\subsection{Training with Unlabeled Data} 
We provide a solution to utilize additional unlabeled images to boost captioning performance.  
Specifically, we use sequence-level knowledge distillation (\textbf{KD}) \cite{kim2016sequence} strategy, where 
the captions produced by a pre-trained autoregressive Transformer teacher model is considered as pseudo target captions for unlabeled images. 
Following previous works on NAT \cite{gu2017non}, 
we also use this KD strategy to generate pseudo target captions for labeled images. 

Before starting CMAL training, we first pre-train the NAIC model with the XE loss (Equation~\ref{eqn:xe}), 
during which we use both the labeled and unlabeled images and their corresponding \textit{pseudo} captions as training data. 
Then during CMAL training (Equation~\ref{eqn:pg-cf}), we use the labeled images and their \textit{real} captions from the original dataset.  
There are two advantages of using real captions instead of pseudo captions for CMAL training: 
first, the reward computation at training time is consistent with the evaluation metric computation at test time,  
\ie the generated caption is compared against the real captions;  
second, unlike previous works on NAT, 
the performance of our method will not be limited by that of the KD teacher model.

\begin{table*}[htbp]
	\centering
	\resizebox{0.85\textwidth}{!}{
		\begin{tabular}{l|cccccc|cc}
			\toprule
			Models & \multicolumn{1}{c}{BLEU-1} & \multicolumn{1}{c}{BLEU-4} & \multicolumn{1}{c}{METEOR} & \multicolumn{1}{c}{ROUGE} & \multicolumn{1}{c}{SPICE} & \multicolumn{1}{c}{CIDEr} & \multicolumn{1}{|c}{Latency} & \multicolumn{1}{c}{Speedup} \\
			\midrule
			\midrule
			\multicolumn{9}{l}{\textbf{Autoregressive models}} \\
			\midrule
			NIC-v2 \cite{vinyals2017show} & / & 32.1  & 25.7  & / & / & 99.8  & / & / \\
			Up-Down \cite{anderson2017bottom} & 79.8  & 36.3  & 27.7  & 56.9  & 21.4  & 120.1 & / & / \\
			VSUA \cite{guo2019vsua} & / & 38.4 & 28.5 & 58.4&  22.0 & 128.6 & / & / \\
			ETA$^\dag$ \cite{li2019entangled} & \bf81.5 & \bf39.3 & 28.8 & \bf58.9 & 22.7 & 126.6  & / & / \\
			ORT$^\dag$ \cite{herdade2019image}   & 80.5  & 38.6  & 28.7  & 58.4  & 22.6  & 128.3 & / & / \\
			AIC$^\dag$ ($\text{bw}=1$) &  79.8&	38.4&	29.0 &	58.7	& 22.8&	126.6  &   134ms    &  1.66$\times$ \\
			AIC$^\dag$ ($\text{bw}=3$) &  80.3&	38.9&	\bf29.1 &	\bf58.9	& \bf22.9&	\bf128.8  &   222ms    &  1.00$\times$ \\
			\midrule 
			\multicolumn{9}{l}{\textbf{Non-autoregressive models}} \\
			\midrule
			MNIC$^\dag$ \cite{gao2019masked} & 75.4  & 30.9  & 27.5  & 55.6  & 21.0    & 108.1 & 61ms & 2.80$\times$ \\
			FNIC $^\dag$\cite{fei2019fast} 						  & /     & 36.2  & 27.1  & 55.3  &20.2     &115.7  & /   &  8.15$\times$ \\
			\midrule
			\multicolumn{9}{l}{\textbf{Non-autoregressive models (Ours)}} \\
			\midrule
			NAIC-base$^\dag$   &   60.7	& 15.9&	18.2&	45.9&	11.9	& 60.6   &    \multirow{5}[1]{*}{\bf16ms}  &  \multirow{5}[1]{*}{\bf13.90$\times$} \\
			\ \ \ \ + weight-init   &  62.3 &	17.1	 &19.0 &	46.8 &	12.6 &	64.6  &       &  \\
			\ \ \ \ + KD &  78.5&	35.3	&27.3 &	56.9	& 20.8&	115.5   &       &  \\
			\ \ \ \ + CMAL &  80.3 &	37.3&	28.1	&58.0 & 	21.8	& 124.0  &       &  \\
			\ \ \ \ + unlabel  &    \bf80.5 &	\bf38.0 &	\bf28.3 &	\bf58.2	& \bf22.0 &	\bf125.5   &       &  \\
			\bottomrule
		\end{tabular}%
	}
	\caption{Generation quality, latency, and speedup on MSCOCO dataset. 
		``$\dag$" indicates the model is based on Transformer architecture. 
		AIC is our implementation of the Transformer-based autoregressive model, which has the same structure as NAIC models and is used as the teacher model for KD. 
		``/" denotes that the results are not reported. 
		``bw" denotes the beam width used for beam search. 
		Latency is the time to decode a single image without minibatching, 
		averaged over the whole test split, and is tested on a GeForce GTX 1080 Ti GPU. 	
		The latency and speedup values of MNIC and FNIC are from the paper. 
	}
	\label{tab:main}%
\end{table*}%

\section{Experiments}
\subsection{Experimental Settings} 
\paragraph{MSCOCO dataset.} MSCOCO \cite{chen2015microsoft} 
is the most popular benchmark for image captioning. 
We use the `Karpathy' splits \cite{karpathy2015deep} that have been used extensively for 
reporting results in prior works. This split contains 113,287
training images with 5 captions each, and 5,000 images for validation and test splits, respectively. 
The vocabulary size is 9,487 words. 
We use the officially released MSCOCO unlabeled images as unlabeled data. 
To be consistent with previous works, we pre-extract image features for all the images following \cite{anderson2017bottom}.

\paragraph{Evaluation metrics.}
We use standard automatic evaluation metrics to evaluate the quality of captions, including BLEU-1/4, METEOR, ROUGE, SPICE, and CIDEr \cite{chen2015microsoft}, 
denoted as B1/4, M, R, S, and C, respectively.

\paragraph{Implementation details.}
Both our NAIC and AIC models closely follow the same model hyper-parameters as Transformer-\textit{Base} \cite{vaswani2017attention} model.  
Specifically, the number of stacked blocks $L$ is 6. 
The AIC model is trained first with XE loss and then with SCST \cite{Rennie2016Self}. 
Beam search with a beam width of 3 is used during decoding of AIC model. 
Our best NAIC model is trained according to the following process. 
We first initialize the weights of NAIC model with the pre-trained AIC teacher model. 
We then pre-train NAIC model with XE loss for 30 epochs. 
During this stage, we use a warm-up learning rate of 
$min( t\times 10^{-4}; 3\times 10^{-4})$, where $t$ is the current epoch number that starts at 1. 
After 6 epochs, the learning rate is decayed by 0.5 every 3 epochs. 
After that, we run CMAL training to optimize the CIDEr metric for about 70 epochs. 
At this training stage, we use an initial learning rate of $7.5\times 10^{-5}$ and decay it by 0.8 every 10 epochs.
Both training stages use Adam \cite{kingma2014adam} optimizer with a batch size of 50. 
By default, we use $k=2$ top-ranking words in CMAL, and use $100,000$ unlabeled images for training. 
We use a fixed number of $N=16$ agents because most of the captions are no longer than this length.

\subsection{Results and Analysis}
\paragraph{General comparisons.} 
We first compare the performance of our methods against other non-autoregressive models and state-of-the-art autoregressive models. 
Among the autoregressive models, ETA, ORT, MNIC, FNIC, and AIC are based on similar Transformer architecture as ours, while others are based on LSTM \cite{hochreiter1997long}. 
MNIC and FNIC are published non-autoregressive image captioning models. 
MNIC adopts an iterative refinement strategy, while FNIC orders words detected in the image with an RNN. 
As shown in Table~\ref{tab:main}, our best model (the last row) 
achieves significant improvements over the previous non-autoregressive models across all metrics,
strikingly narrowing their performance gap between AIC from 13.1 CIDEr points down to only 3.3 CIDEr points. 
Furthermore, we achieve comparable performance with state-of-the-art autoregressive models.  
Comparing speedups, our method obtains a significant speedup of more than a factor of 10 over the 
autoregressive counterpart, with latency\footnote{The time for image feature extraction is not included in latency. } reduced to about 16ms. 
We show the results of online MSCOCO evaluation in Table~\ref{tab:server}.
	
\begin{table*}[tbp]
	\centering
	\resizebox{\textwidth}{!}{
		\begin{tabular}{@{\extracolsep{3pt}}@{\kern\tabcolsep}lrrrrrrrrrrrrrr}
			\toprule
			\multicolumn{1}{c}{\multirow{2}[4]{*}{Model}}  & \multicolumn{2}{c}{BLEU-1} & \multicolumn{2}{c}{BLEU-2} & \multicolumn{2}{c}{BLEU-3} & \multicolumn{2}{c}{BLEU-4} & \multicolumn{2}{c}{METEOR} & \multicolumn{2}{c}{ROUGE-L} & \multicolumn{2}{c}{CIDEr-D} \\
			\cmidrule{2-3}  \cmidrule{4-5} \cmidrule{6-7} \cmidrule{8-9} \cmidrule{10-11} \cmidrule{12-13} \cmidrule{14-15} %
			& \multicolumn{1}{c}{c5} & \multicolumn{1}{c}{c40} & \multicolumn{1}{c}{c5} & \multicolumn{1}{c}{c40} & \multicolumn{1}{c}{c5} & \multicolumn{1}{c}{c40} & \multicolumn{1}{c}{c5} & \multicolumn{1}{c}{c40} & \multicolumn{1}{c}{c5} & \multicolumn{1}{c}{c40} & \multicolumn{1}{c}{c5} & \multicolumn{1}{c}{c40} & \multicolumn{1}{c}{c5} & \multicolumn{1}{c}{c40} \\
			\midrule
			Up-Down$^*$ \cite{anderson2017bottom} & 80.2  & 95.2  & 64.1  & 88.8  & 49.1  & 79.4  & 36.9  & 68.5  & 27.6  & 36.7  & 57.1  & 72.4  & 117.9  & 120.5  \\
			VSUA \cite{guo2019vsua} & 79.9  & 94.7  & 64.3  & 88.6  & 49.5  & 79.3  & 37.4  & 68.3  & 28.2  & 37.1  & 57.9  & 72.8  & 123.1  & 125.5  \\
			ETA$^*$ \cite{li2019entangled} & 81.2& 95.0 &65.5 &89.0& 50.9& 80.4& 38.9& 70.2& 28.6 &38.0& 58.6 &73.9& 122.1& 124.4 \\
			\midrule
			NAIC-CMAL (Ours) & 79.8 & 94.3 & 63.8 & 87.2 & 48.8 & 77.2 & 36.8 & 66.1 & 27.9 & 36.4 & 57.6 & 72.0 & 119.3 &  121.2 \\
			\bottomrule
		\end{tabular}%
	}
	\caption{Results on the online MSCOCO test server. $*$ denotes ensemble model.  }
	\label{tab:server}%
\end{table*}%

\paragraph{Ablation study.}
We conduct an extensive ablation study with the proposed NAIC model. 
The results are shown in the bottom of Table~\ref{tab:main}, 
where ``NAIC-base" is the naive NAIC model trained from scratch using XE loss, 
``KD" represents using knowledge distillation with AIC as the teacher model, 
``CMAL" denotes further applying our proposed CMAL algorithm for CIDEr optimization, 
``unlabel" means using additional 100,000 unlabeled data during XE training,  
and ``weight-init" denotes initializing the weights of NAIC with AIC model. 
We specially consider the case when not using weight-init  
because it may not be possible to find an autoregressive model that has the same structure 
as a novelly designed non-autoregressive model. 
We have the following observations. 
First, initializing NAIC model's weights with its pre-trained AIC can consistently improve the performance.  
Second, NAIC-base performs extremely poorly compared to AIC. 
Third, we see that training on the distillation data during XE training improves the CIDEr score to 115.5. 
However, there still remains a large performance gap between this model and the AIC teacher. 
Fourth, applying our CMAL training on top of the above XE trained model 
significantly improves the performance by 8.5 CIDEr points.  
Last, using additional unlabeled data for training further boosts the performance by 1.5 CIDEr points. 
\begin{table}[tp]
	\centering
	\resizebox{0.43\textwidth}{!}{
		\begin{tabular}{lrrrrrr}
			\toprule
			\multicolumn{1}{l}{Baseline $b$}  & \multicolumn{1}{c}{B1} & \multicolumn{1}{c}{B4} & \multicolumn{1}{c}{M} & \multicolumn{1}{c}{R} & \multicolumn{1}{c}{S} & \multicolumn{1}{c}{C} \\
			\midrule
			\midrule
			\multicolumn{1}{l}{\textbf{w/o weight-init:}} \\
			\ \ \ \ XE & 77.7 & 34.8 &	26.9 &	56.3 &	20.3 &	113.9 \\
			\ \ \ \ None & 65.6& 	19.4 &	22.7 &	48.9 &	15.8 &	91.4 \\
			\ \ \ \ MA & 75.6 &	28.7 &	24.4 &	53.6 &	17.9 &	103.3  \\
			\ \ \ \ SC & 79.0 	&34.6 &	26.9 &	56.2 &	20.6 &	118.1 \\
			\ \ \ \ CF & \bf79.9 &	\bf36.5 &	\bf27.7 &	\bf57.4 &	\bf21.4 &	\bf122.1 \\
			\midrule
			\multicolumn{1}{l}{\textbf{w/ weight-init:}} \\
			\ \ \ \ XE & 78.5 &	35.3 &	27.3 &	56.9 &	20.8 &	115.5 \\
			\ \ \ \ None & 78.6 &	33.7 &	26.5 &	56.1 &	20.2& 	115.2 \\
			\ \ \ \ MA & 79.0 &	34.1 	&26.6 	&56.3 	&20.2 &	116.1 \\
			\ \ \ \ SC & 79.6 	&36.5 &	27.6 &	57.4 	&21.4 &	121.2 \\
			\ \ \ \ CF & \bf80.3 &	\bf37.3 &	\bf28.1 	& \bf58.0 	&\bf21.8 &	\bf124.0 \\
			\bottomrule
		\end{tabular}%
	}
	\caption{Comparison of using various baselines $b$ in Equation \ref{eqn:pg}.  
		XE: the performance after pre-training with cross-entropy loss.}
	\label{tab:baseline}%
\end{table}%

\begin{table}[tp]
	\centering
	\resizebox{0.38\textwidth}{!}{
		\begin{tabular}{crrrrrr}
			\toprule
			\multicolumn{1}{c}{top-$k$} & \multicolumn{1}{c}{B1} & \multicolumn{1}{c}{B4} & \multicolumn{1}{c}{M} & \multicolumn{1}{c}{R} & \multicolumn{1}{c}{S} & \multicolumn{1}{c}{C} \\
			\midrule
			\midrule
			1 & 80.1 &	\bf37.4 &	28.0 &	57.9 &	21.7 &	123.7 \\
			2 & \bf80.3 &	 37.3 &	\bf28.1 	& \bf58.0 	&\bf21.8 &	\bf124.0 \\
			5 & 80.1 &	37.3 &	28.0 &	58.0 &	21.7 &	123.7 \\
			\bottomrule
		\end{tabular}%
	}
	\caption{Effect of top-$k$ size in CMAL. }
	\label{tab:topk}%
\end{table}%

\paragraph{Comparison of various reward baselines $b$.} 
To evaluate the effectiveness of our counterfactual (CF) baseline, 
we compare it with two widely-used baselines in policy gradient, 
\ie Moving Average \cite{weaver2001optimal} (MA) and Self-Critical \cite{Rennie2016Self} (SC), and not using a baseline (None), \ie$b=0$. 
MA baseline is the accumulated sum of the previous rewards with exponential decay. 
SC baseline is the received reward when all agents directly take greedy actions. 
As shown in Table~\ref{tab:baseline}, our CF baseline consistently outperforms all the other compared methods. 
Noteworthy that the performance gaps between our CF baseline and other baselines become larger when 
trainings start from a poor-performed model (\ie XE model w/o weight-init). 
That is, our method is less sensitive to model initialization, 
suggesting its ability to enable more robust and stable reinforcement learning.  
None and MA severely degrades the performance compared to XE model when not using weight-init, 
but they perform similar to XE model when using weight-init. 
While SC considerably outperforms XE model, it is still inferior to CF. 
The reason is that both MA and SC are agent-agnostic global baselines, 
which cannot address the multi-agent credit assignment problem, 
while our CF baseline is agent-specific.

\paragraph{Effect of top-$k$ size.}
As shown in Table~\ref{tab:topk}, the model is not sensitive to the choice of top-$k$ size.  
Using a small $k$ of 2 could achieve fairly good performance.

\paragraph{Number of unlabeled images.} 
In Table~\ref{tab:unlabel}, we show the results after XE and CMAL training 
when using 0, 50,000 and 100,000 unlabeled images respectively. 
Generally, using more unlabeled images could lead to better performance.  
XE training benefits more from the unlabeled images than CMAL training because 
we directly use the unlabeled images during XE training while not using them for CMAL.

\begin{table}[tp]
	\centering
	\resizebox{0.43\textwidth}{!}{
		\begin{tabular}{cc|rrrrrr}
			\toprule
			\#unlabel & stage & \multicolumn{1}{c}{B1} & \multicolumn{1}{c}{B4} & \multicolumn{1}{c}{M} & \multicolumn{1}{c}{R} & \multicolumn{1}{c}{S} & \multicolumn{1}{c}{C} \\
			\midrule
			\midrule
			\multirow{2}[0]{*}{0} & XE & 78.5 &	35.3 &	27.3 &	56.9 	&20.8& 	115.5  \\
			& CMAL &  80.3 	&37.3 &	28.1 &	58.0& 	21.8 &	124.0  \\
			\midrule
			\multirow{2}[0]{*}{50k} & XE & 78.8 	&36.2 &	27.6 &	57.2 &	21.1 	&118.1  \\
			& CMAL &   80.2 & 	37.6 & 	28.1 & 	58.1 & 	21.9 & 	124.8 \\
			\midrule
			\multirow{2}[0]{*}{100k} & XE &  79.0 &	36.2 &	27.7 &	57.3 &	21.2& 	118.3 \\
			& CMAL &   \bf80.5 &	\bf38.0 &	\bf28.3 &	\bf58.2 &	\bf22.0 &	\bf125.5 \\ 
			\bottomrule
		\end{tabular}%
	}
	\caption{The results after XE and CMAL training when using different numbers of unlabeled images.}
	\label{tab:unlabel}%
\end{table}%

\begin{figure}[!t] 
	\centering
	\includegraphics[width=3.4in]{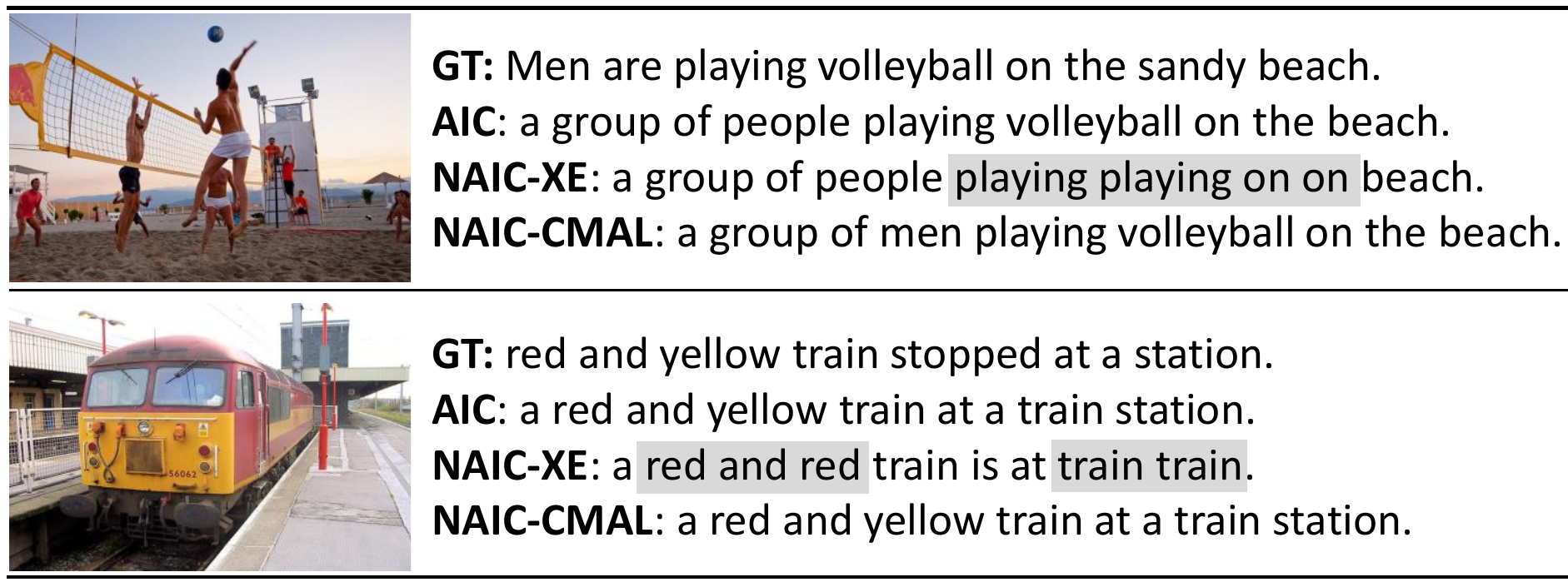}
	\caption{
		Two examples of the generated captions. 
		GT is a ground-truth caption. NAIC-XE and NAIC-CMAL are our NAIC model after XE and CMAL training, respectively. 
		Repeated words are highlighted in gray.	
	}
	\label{fig:example}
\end{figure}

\paragraph{Qualitative analysis.}
We present two examples of generated image captions in Figure~\ref{fig:example}. 
As can be seen, repeated words and incomplete content are most prevalent in the XE trained NAIC model, 
showing that the word-level XE training often results in decoding inconsistency problem. 
With our CMAL training, the sentences become far more consistent and fluent. 

\section{Conclusion} 
We have proposed a non-autoregressive image captioning model and a novel counterfactuals-critical multi-agent learning algorithm. 
The decoding inconsistency problem in non-autoregressive models 
is well addressed by the combined effect of the cooperative agents, sentence-level team-reward, and agent-specific counterfactual baseline. 
The caption quality is further boosted by using unlabeled images.  
Results on MSCOCO image captioning benchmark show that our non-autoregressive model can achieve a performance comparable to state-of-the-art autoregressive counterparts, 
while at the same time enjoy $13.9\times$ inference speedup. 

\paragraph{Acknowledgments.}
This work was supported by Beijing Natural Science Foundation (No.4192059) and National Natural Science Foundation of China (No.61922086, No.61872366, and No.61872364). 

\bibliographystyle{named}
\bibliography{references}

\end{document}